\documentclass{article}

 \PassOptionsToPackage{numbers, compress}{natbib}
\usepackage[preprint]{nips_2018}

\usepackage[utf8]{inputenc} % allow utf-8 input
\usepackage[T1]{fontenc}    % use 8-bit T1 fonts
\usepackage{hyperref}       % hyperlinks
\usepackage{url}            % simple URL typesetting
\usepackage{booktabs}       % professional-quality tables
\usepackage{amsfonts}       % blackboard math symbols
\usepackage{nicefrac}       % compact symbols for 1/2, etc.
\usepackage{microtype}      % microtypography
\usepackage{graphicx}
\usepackage{bm}
\usepackage{float}
\usepackage{subfigure}
\usepackage{verbatim}
\newcommand{\etal}{et al.}
\newcommand{\Rspace}{\mathcal{R}}

\usepackage[x11names]{xcolor}
\usepackage{comment}
\usepackage{etoolbox}
\usepackage{soul}
\title{DiDA: Disentangled Synthesis for Domain Adaptation}

\author{
  Jinming Cao\\
  Shandong University\\
  \And
  Oren Katzir \\
  Tel Aviv University \\
  \And
  Peng Jiang \\
  Shandong University \\
  \And
  Dani Lischinski \\
  The Hebrew University of Jerusalem \\
  \And
  Danny Cohen-Or \\
  Tel Aviv University \\
  \And
  Changhe Tu  \\
  Shandong University \\
  \And
  Yangyan Li \\
  Shandong University \\
}

\begin{document}

\maketitle

\begin{abstract}
	
Unsupervised domain adaptation aims at learning a shared model for two related, but not identical, domains by leveraging supervision from a \emph{source} domain to an unsupervised \emph{target} domain.
A number of effective domain adaptation approaches rely on the ability to extract discriminative, yet domain-invariant, latent factors which are common to both domains.
% Not sure the next sentence is necessary, the abstract night read better without it... (Dani)
Extracting latent commonality is also useful for disentanglement analysis, enabling separation between the common and the domain-specific features of both domains.
In this paper, we present a method for boosting domain adaptation performance by leveraging disentanglement analysis.
The key idea is that by learning to separately extract both the common and the domain-specific features, one can synthesize more target domain data with supervision, thereby boosting the domain adaptation performance. 
Better common feature extraction, in turn, helps further improve the disentanglement analysis and disentangled synthesis.
We show that iterating between domain adaptation and disentanglement analysis can consistently improve each other on several unsupervised domain adaptation tasks, for various domain adaptation backbone models.

\end{abstract}

\section{Introduction}
\label{sec:introduction}

Many machine learning solutions are supervised, requiring well-annotated training datasets.
Such annotations can be overly expensive to attain for the amount of data required for plausible performance by today's deep neural networks.
Domain adaptation approaches attempt to compensate for lack of annotated data in a \emph{target domain}, by adapting information from a \emph{source domain}, for which annotated data is easier to obtain.
For example, in many cases, it is easy to generate synthetic data (source domain) with inherent annotations, and use this data for the supervised training of a network that is ultimately intended for carrying out a task over real data (the target domain).
While the source data may closely resemble the real target data, there are typically inevitable differences between the two domains, and domain adaptation faces the challenge of overcoming such \emph{domain shifts}.

In this work, we target the unsupervised domain adaptation scenario, where the source domain data, either synthetic or real, is fully annotated, while the target domain data has no annotations whatsoever. 
Existing approaches to domain adaptation, which are reviewed in more detail in the next section, can be broadly classified into two categories: methods based on domain transfer (e.g., by translating data from one domain to another), and those based on embedding both domains in a common feature space.
The goal of the current work is to propose a framework which improves the performance of methods in the latter category.

In domain adaptation, it is typically assumed that there is a significant commonality between the source and the target domains in aspects relevant to the task at hand~\cite{Ben-David2010} (e.g., classification).
Thus, data items in each of the two domains can be conceptually factored into their task-relevant
\emph{common features} and their \emph{domain-specific features}.
Domain adaptation could thus be achieved by training a model on the common features extracted from the annotated source domain, and applying it on the common features extracted from the target domain at test time.

However, neither the common features, nor the domain-specific ones are given explicitly.
Rather, these features are typically latent, and thus it is necessary to train a model to extract the common features, regardless of the domain of the input, in order to obtain good performance on target domain data.  

Several existing methods attempt to embed both domains in a shared feature space, which effectively amounts to extracting their common features. 
In this paper, we present a framework for boosting the performance of such methods by iterative interleaving of domain adaptation and disentanglement analysis.
The key idea is that rather than focusing solely on extracting the common feature, we attempt to extract a disentangled latent representation of the data from both domains, i.e., to explicitly obtain its common and specific features.

The ability to disentangle both the source and the target domain data paves the way for attribute swapping~\cite{lample2017fader}, or \emph{disentangled synthesis}~\cite{hadad2017two,huang2018munit}, which enables the generation of additional annotated synthetic training data.
Specifically, this data is generated by combining the common features and the corresponding annotations of source domain data with domain-specific features extracted from the target domain data.
The resulting annotated synthesized data is closer to the target domain, and may be used to improve the common feature extraction, thereby boosting domain adaptation performance.
Better common feature extraction, in turn, helps further improve the disentanglement analysis and disentangled synthesis.

Following a review of related work in the next section, we describe our framework in more detail in Section \ref{sec:method}.
Our experiments in Section \ref{sec:experiments} show that by iterating between domain adaptation and disentanglement analysis, as outlined above, the performance of domain adaptation can be significantly improved.
We demonstrate this on several unsupervised domain adaptation benchmarks, using different domain adaptation backbone methods.

\section{Related Work}\label{sec:related_work}
%In this section, we will give a brief review about the two models that our method depends on.

\subsection{Domain Adaptation}\label{sec:domain_adaptation}

Many works have addressed the problem of domain adaptation over the years. Here, we mainly focus on methods that are based on deep neural networks, as these represent the state-of-the-art and are the most relevant to our scheme.

The problem of domain adaptation was theoretically studied by Ben-David et al.~\cite{Ben-David2010}, who suggest that a good domain adaptation method should assimilate the common features of source and target domains, while disassimilating the domain specific features.
Following this principle, two branches of domain adaptation methods have emerged, consisting of methods based on domain transfer~\cite{Bousmalis2017CVPR,Hoffman2017CyCADA,russo2017source,saito2017asymmetric,Tzeng2017CVPR} and methods based on common feature embedding~\cite{ganin2016domain,ghifary2016deep,haeusser2017associative,long2015learning,long2016deep,sankaranarayanan2017generate,tzeng2014deep}.
%Among them, adversarial networks~\cite{goodfellow2014generative} is the most common approach used to do mapping between images or features, one major problem by this fashion is the sensitivity to the number of labeled data, that when samples of source domain are much less, they will not provide sufficient supervision.

Domain transfer based methods generalize the functional component from the source domain to the target. Bousmalis et al.~\cite{Bousmalis2017CVPR} use a generative adversarial network (GAN \cite{goodfellow2014generative}) to translate source images to target images and assign them their corresponding source labels. Conversely, Tzeng et al.~\cite{Tzeng2017CVPR} transfer the target domain feature space to that of the source domain in an adversarial manner, such that features from the two domains can not be distinguished. Further, Russo et al.~\cite{russo2017source} and Hoffman et al.~\cite{Hoffman2017CyCADA} ensure mutual transfer between the source and the target datasets by adding cycle consistency terms. Saito et al.~\cite{saito2017asymmetric} train two classifiers on the source dataset, in order to artificially label the target dataset.

A different approach to domain adaptation is to embed both domains in a shared feature space, which minimizes the discrepancy between the source and the target data. Several methods \cite{long2015learning,long2016deep,sun2016deep,tzeng2014deep} use hand-crafted metrics, such as Maximum Mean Discrepancy (MMD)~\cite{gretton2012kernel} and second-order statistics
(covariances). Other methods learn a discrepancy metric. For example, Ganin \etal~\cite{ganin2016domain} propose Domain Adversarial Neural Networks (DANN) which employ the same encoder network, followed by classification layers, to correctly classify the source dataset,
while fooling a domain classifier.
%Bearing similar concept, \cite{Tzeng2017CVPR}, first learn an embedding in order to classify only the source domain, and then train a target-encoder in adversarial manner against a source-features discriminator.
Haeusser \etal~\cite{haeusser2017associative} produce statistically domain invariant embeddings by reinforcing associations between source and target data directly in embedding space.
A slightly different approach is presented by Ghifary et al.~\cite{ghifary2016deep}, where common feature assimilation is achieved implicitly by the ability of a decoder to reconstruct the input source and target images. In a similar spirit, \cite{sankaranarayanan2017generate} requires a generator from the encoded features to be able to generate samples which follow the same distribution as the source dataset.

%Bousmalis\etal~\cite{bousmalis2016domain} combines several concepts from previous works. Three encoders are trained simultaneously, one for the shared embedding and two for each dataset unique characteristics. The shared embedding loss combines both hand-crafted and learnt loss as well as a classification and reconstruction losses.}

\subsection{Disentangled Synthesis}\label{sec:disentangled_synthesis}

Deep neural networks often give rise to latent feature spaces where multiple hidden factors of variation in the data are highly entangled.  
Thus, disentanglement is required for latent feature manipulation, such as data synthesis and attribute transfer. 
Disentanglement is often carried out using various autoencoders, trained in different manners, depending on the problem setting.

When the entire data is labeled, latent features that code other auxiliary information can be extracted by adding an adversarial component~\cite{lample2017fader} or via variational inference~\cite{KingmaNIPS2014},
ensuring that auxiliary latent features can not be used to predict the label.
Thus, auxiliary latent features and given labels are complementary, in the sense that the auxiliary latent features do not contain any label information, but can be used together with the labels to reconstruct the input data.
However, these approaches do not support novel labels at test time, i.e., they are limited to the set of labels seen during training, which is undesirable for certain applications.
%\DaniCom{not clear what ``do not support novel labels at test time'' means? And why is it so?}. \JumpCom{These works do attribute transfer, here label actually means attribute, (I remember I used "label(attribute)" here). Given the attribute of data and with auxiliary latent features (self or others), these works can synthesis data with identical attribute but different style, such as facial expression transfer. But these work can not handle the case given data with no attribute provided, because they do not have classifier and usually these attribute is also hard for users to describe. This statement is from "hadad2017two".}

Hadad et al.~\cite{hadad2017two} address this limitation by introducing a two-step disentanglement approach. The first step trains an encoder for extracting the task related features and a classifier for predicting their label. The second step performs the disentanglement, by training another encoder using an adversarial classifier to extract the complementary features. While labels are used during training, they are not required for the extraction of disentangled features at test time.

%however, features from two encoders may not be mutually exclusive anymore \DaniCom{also not clear, why not? Maybe I don't understand it because I don't understand what mutually exclusive means here} \JumpCom{This paper firstly train a encoder-classifier to predict label, they claim the feature from encoder only contains information of labels, and use these feature to do data synthesis as we do, but in our previous experiments, these feature do contains other information such as style.}.
%which  makes it no longer limited to the labels seen during training. 

When given data from two correlated domains, it is possible to synthesize new data by combining parts of features from both domains and ensuring that reconstructed data is indistinguishable from real data by an adversarial objective.
In this way, latent features of both domains are effectively disentangled
%\DaniCom{not clear why the above achieves disentanglement...}
%\JumpCom{If synthesis data by combining and swapping of features from two domains can fool the discriminator well, then these feature should be very clean and not disentangled. Otherwise, synthesis data will have many artifact.}
without need of any explicit label information~\cite{huang2018munit,mathieu2016disentangling}.
However, in this case, the features are not necessarily disentangled into common and domain-specific parts, which is what we need in order to boost common feature extraction for domain adaptation.

In this work, we have two correlated domains, source and target, with labels available for the source domain only.
Our approach is related to Bousmalis et al.~\cite{bousmalis2016domain}, who perform disentanglement by extracting common features using a domain adaptation method~\cite{ganin2016domain}, while forcing domain-specific features to be orthogonal to common features.
%However, some people[] argue \DaniCom{ok, if we want to claim that we really need to provide a specific reference. Do we have one?} \JumpCom{Sorry, I forget where I met this statement. Does anyone know?} that orthogonal features are not necessarily mutually exclusive.
However, in our work, we adopt a two step scheme inspired by Hadad et al.~\cite{hadad2017two}, where we replace their first encoding and classification step with adversarial domain adaptation~\cite{ganin2016domain}.
%From the theoretical analysis in the next section, we believe our method could disentangle features to mutually exclusive. 

While we are not the first to use domain adaptation in conjunction with domain disentanglement, previous work~\cite{bousmalis2016domain} only used disentanglement to assist in the extraction of a common feature space, thereby allowing domain adaptation by using a classifier trained on the common features. In contrast, we interleave domain adaptation and disentanglement and use disentangled synthesis to generate synthetic annotated target domain data, which is then fed back into the domain adaptation step.

\section{Method}
\label{sec:method}

The main idea of our approach is to alternate between domain adaptation and disentanglement analysis.
By alternating between these two tasks, each task is able to leverage the other to achieve better performance, the end result being improved domain adaptation.

\begin{figure*}[t!]
	\centering
	\centering
	\includegraphics[width=0.9\textwidth]{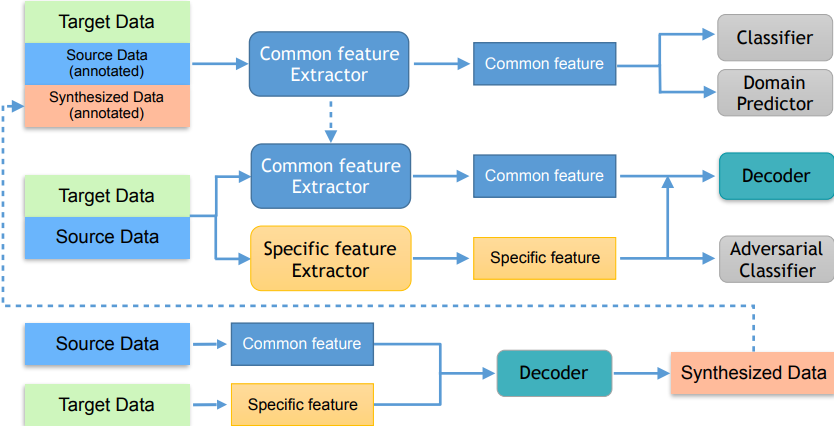}
	\caption{Our iterative Domain Adaption + Disentanglement method. Top: domain adaptation stage; Middle: disentanglement analysis stage; Bottom: disentangled synthesis.}
	\label{fig:pipeline}
\end{figure*}

More specifically, the training iterates over two consecutive stages.
The first, \emph{domain adaptation} stage, learns to extract (from both the source domain and the target domain data) a \emph{common feature}, which is discriminative, yet domain invariant.
In other words, the common feature should ideally capture all the information relevant to the classification task, but none of the information that is domain specific.
Our approach can be applied to any domain adaption method which learns such \emph{common feature} representation, as illustrated in the top row of the diagram in Figure~\ref{fig:pipeline}.
The common feature extractor is optimized according to the chosen method, usually via a combination of a classification loss (computed only over the annotated training examples of source domain), and a domain loss, applied either in adversarial fashion or using a closed form metric between source and target features.

The second, \emph{disentanglement} stage, utilizes the trained common feature extractor from the previous stage and trains a \emph{domain-specific feature extractor}.
Following the method proposed by Hadad et al.~\cite{hadad2017two}, the goals of this stage are that the combination of the two features suffices to reconstruct the input, while the specific feature is category agnostic.
The former is achieved using a reconstruction loss, while the latter using an adversarial classifier, as depicted in the second row of Figure~\ref{fig:pipeline}.

\begin{figure*}[t!]
	\centering
	\centering
	\includegraphics[width=0.9\textwidth]{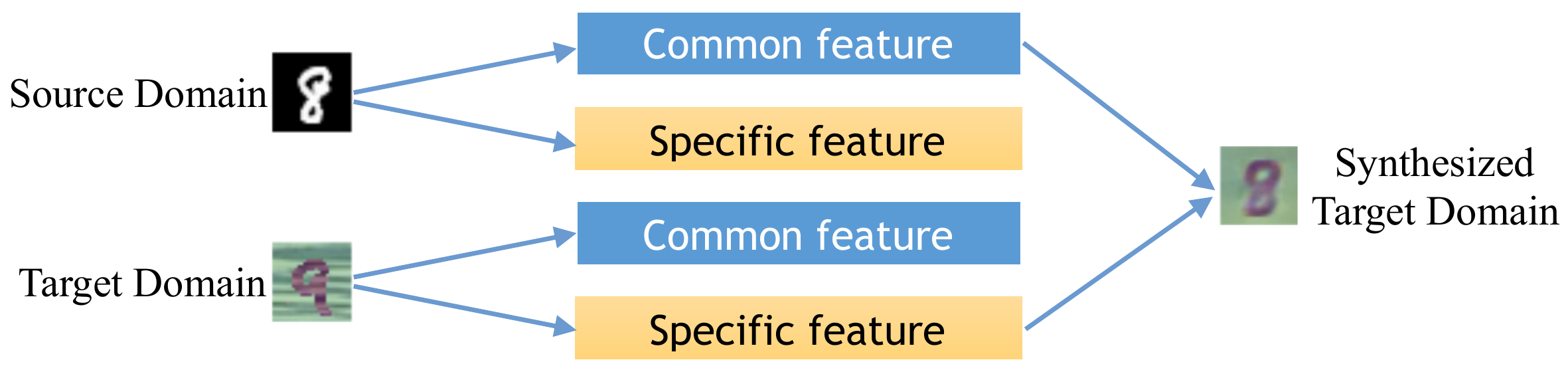}
	\caption{Disentangled synthesis for generation of annotated samples in a synthesized target domain. The example shown is for the MNIST to MNISTM adaptation.}
	\label{fig:hybrid_img}
\end{figure*}

The ability to extract common and domain-specific features makes it possible to generate new annotated data, by combining the label and the common feature of source domain training examples with domain-specific features extracted from target domain samples, and reconstructing (bottom row of Figure~\ref{fig:pipeline}).
We refer to this process as \emph{attribute swapping} or \emph{disentangled synthesis}, and demonstrate it in Figure~\ref{fig:hybrid_img}.
The annotated source domain training data of the first stage is thereby augmented with synthesized annotated data, which ideally looks like the unlabeled data from the target domain.
Armed with this augmented data, we can iterate the entire process with the hope that both the domain adaptation and the disentanglement are improved by each iteration.

In the remainder of this section, we provide more details regarding each of the steps outlines above.

\paragraph{Domain Adaptation.}

As mentioned earlier, in this work we address the unsupervised domain adaptation scenario, where the training data consists of $N_s$ labeled samples from a source domain,
$\bm{X}_s=\left\{\left(x_i^s, y_i^s\right)\right\}_{i=0}^{N_s}$,
along with $N_t$ unlabeled samples from a target domain,
$\bm{X}_t=\left\{\left(x_i^t\right)\right\} _{i=0}^{N_t}$.
The first stage of our approach can employ any domain adaptation method that learns a mapping of the above training samples to a common latent space.
A number of such methods have been reviewed in Section \ref{sec:domain_adaptation}.
%, e.g., 
%\cite{ganin2016domain,haeusser2017associative,sun2016deep,tzeng2014deep}
These methods typically employ a loss with two main components, namely, classification loss and domain loss:
$\mathcal{L}_{\mathit{DA}} = \mathcal{L}_{\mathit{class}} + \alpha \mathcal{L}_{\mathit{domain}},
$
where $\alpha \in \Rspace$ is the assimilation weight. 

The classification loss is shared by all domain adaption methods and is usually  the negative log-likelihood of the classifier prediction, given the ground truth labels of samples from the source dataset, i.e., 
$
\mathcal{L}_{\mathit{class}} = -\sum_{i=0}^{N_s} y_i^s \cdot \log \hat{y}_i^s,
$
where $y_i^s, \hat{y}_i^s$ are the label and prediction, respectively, for the sample $x^s_i \in \bm{X}_s$. The domain loss differs from method to method, while the goal remains the same: assimilate the common feature of both source and target domains. This loss can be either hand-crafted \cite{tzeng2014deep, haeusser2017associative, sun2016deep}, or adversarial \cite{ganin2016domain}. In this work we report the results using both kinds of approaches, and we refer the reader to the original papers for more details on the specific loss design. The illustration of the domain adaptation stage in the top row of Figure \ref{fig:pipeline} corresponds to the domain-adversarial training of Ganin et al.~\cite{ganin2016domain}. 

\paragraph{Disentanglement.}
In this stage, we adopt the adversarial training method of Hadad et al.~\cite{hadad2017two}. As explained earlier, we freeze the common feature extractor from the previous stage and train a domain-specific feature extractor.
The domain-specific feature should enable reconstruction (together with the common feature), but it should not enable classification. This is achieved by training the specific feature extractor jointly with a reconstruction \emph{Decoder} and an \emph{Adversarial Classifier}, as illustrated in the middle row of Figure \ref{fig:pipeline}.

The Adversarial Classifier is trained to minimize the classification loss, 
$
\mathcal{L}_\mathit{AClass} = -\sum_{i=0}^{N} y_i \cdot \log \hat{y}_i
$
over all training samples (source and target), where for the target domain samples we use the labels predicted by the classifier trained in the domain adaptation stage.

The Specific feature Extractor and the Decoder are trained together to minimize the disentanglement loss:
$
\mathcal{L}_\mathit{Di} = \mathcal{L}_\mathit{rec} - \beta \mathcal{L}_\mathit{AClass},
$
where $\beta$ is the adversary weight, and $\mathcal{L}_\mathit{rec}$ is the reconstruction loss, computed over samples from both domains:
$
\mathcal{L}_{\mathit{rec}} = {\mathit{MSE}} (x_i, \hat{x}_i^\mathit{rec}),
$
where $x_i, \hat{x}_i^\mathit{rec}$ are the input sample and its reconstruction, respectively. This results in adversarial training, since the adversarial classifier strives to minimize $\mathcal{L}_\mathit{AClass}$, while the extractor strives to maximize it.

\paragraph{Iteration.}
After learning to disentangle the inputs into common and specific features, we can use attribute swapping or disentangled synthesis to generate synthetic labeled data, whose labels and common features come from the source domain, while the specific features come from the target domain, as shown in Figure \ref{fig:hybrid_img}.
The resulting annotated data is used as additional training data for the domain adaptation stage in the next iteration. 
%\JMCom{We think that our formula is relatively simple, there is no need to take another line, so we put it in the text. Because there didn't have enough effective content in Training Detail, I put it in the experimental part.}

\iffalse
\paragraph{Training Detail.}
All our models are implemented using Pytorch. We use MSE for $\mathcal{L}$$_{Re}$, and use cross-entropy loss for $\mathcal{L}$$_{class}$, $\mathcal{L}$$_{domain}$ and $\mathcal{L}$$_{Aclass}$. \DaniCom{Seems repetitive, no? Didn't we say this earlier?}\JMCom{I don't know what other kind of 'Detail' should be write here.} The $\alpha$, $\beta$ for each dataset was chosen independently. We apply the Adam optimization method \cite{kingma2014adam} in the training of Specific feature Extractor and Decoder. For the Common feature Extractor, Classifier, Domain Predictor, and Adversarial Classifier, we use stochastic gradient descent.
\fi

\section{Experiments}
\label{sec:experiments}

Our method is a general framework for boosting domain adaptation. To demonstrate its effectiveness, we implemented various state-of-the-art domain adaptation methods. For the detailed description of the involved network architectures and optimizer choices, please refer to Supp. Material Section~\ref{sec:network_details}.

We first introduce the domain adaptation benchmark datasets, then we show extensive quantitative and qualitative evaluations of our method, demonstrating its boosting of various backbone domain adaptation methods on various benchmark datasets.

\subsection{Domain Adaptation Benchmarks}

\paragraph{MNIST $\rightarrow$ MNISTM.}
Here the labeled source domain is the MNIST dataset \cite{lecun1998gradient} of handwritten digits over a black background. The unlabeled target domain is the MNISTM dataset, where the digits are blended over natural color image patches, using the procedure described by Ganin et al.~\cite{ganin2016domain}. In this case, the features of the digits are the common features domain adaptation methods would like to extract, and the features of the background color image patches are specific to the target domain.
%Due to the presence of the color image patches in the background, a classifier trained solely on MNIST works poorly on MNIST-M.

%Following Ganin et al.~\cite{ganin2016domain}, we use MNIST \cite{lecun1998gradient} as the source domain and generate the MNIST-M dataset as the target domain. Random background patches are extracted from the color photo BSD500 dataset \cite{arbelaez2011contour}, and the absolute value of the difference of each color channel with an MNIST image is taken. The single channel of the MNIST images was replicated three times to match those of the MNIST-M images (RGB), and the size of images are 28 $\times$ 28. This yields a color image, which can be easily identified by a human, but is significantly more difficult for a machine compared to MNIST.

\paragraph{MNIST $\rightarrow$ USPS.}
Here, again, the MNIST dataset is used as the source domain, while images of the same 10 digits from the USPS dataset \cite{denker1989neural} are used as the target domain. In this setting, the ``style'' of the USPS digits, together with the background in USPS, are specific to the target domain. The USPS dataset has a training set of 7291 images of size 16 $\times$ 16. We follow the  protocol described by Long et al.~\cite{long2013transfer} and randomly sample 2000 images from MNIST and 1800 images from USPS.

\paragraph{SVHN $\rightarrow$ MNIST.}
In this setting, Street View House Numbers (SVHN) dataset \cite{netzer2011reading} is used as the source domain and MNIST becomes the target domain. The SVHN dataset contains house number signs extracted from Google Street View. We follow the same procedure described in \cite{haeusser2017associative} for both SVHN and MNIST images, to create images of size $32 \times 32$ pixels. Note that in this case, the domain specific features are quite ``plain'' compared with those in the previous two settings.

\subsection{Quantitative Evaluation}

The goal of our method is for boosting domain adaptation methods, so we evaluate its effectiveness by comparing the accuracies achieved with/without applying our method over various backbone domain adaptation methods on the aforementioned benchmark datasets.

\begin{figure*}[t!]
	\centering
	\centering
	\includegraphics[width=1.0\textwidth]{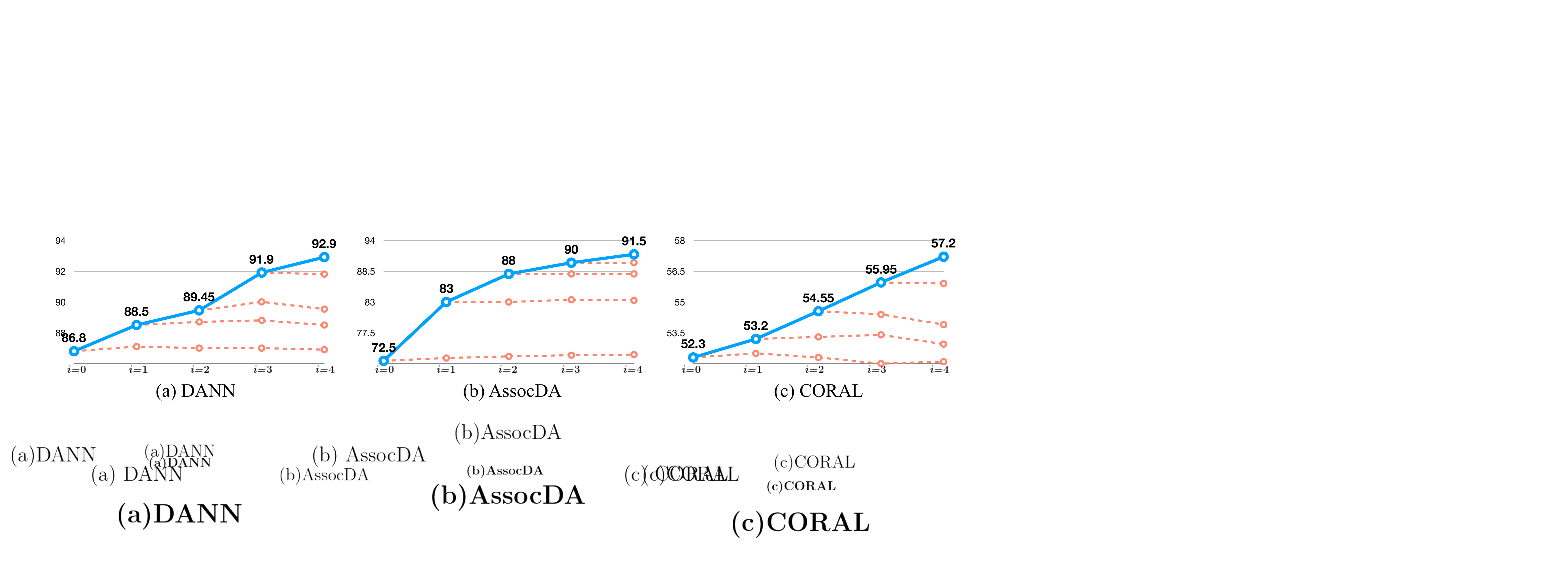}
	\vspace{-0.04\linewidth}
\caption{\label{fig:accvis}
	Classification accuracy (\%) over the DiDA iterations on MNIST $\rightarrow$ MNISTM task, using different domain adaption methods as backbones. The blue circles indicate the accuracy at each DiDA iteration. The orange dots indicate the accuracies achieved by adding the same number of training iterations, but without using new synthetic data (updated by the improved disentanglement achieved by subsequent iterations). This shows that the improvement in accuracy are due to more/better synthetic data, rather than just more training iterations.
}
\end{figure*}

\paragraph{Evaluation on Different DA Methods.} Our method uses disentangled synthesis for boosting domain adaptation, and we evaluated it on three different domain adaption methods: DANN~\cite{ganin2016domain}, AssocDA~\cite{haeusser2017associative} and CORAL~\cite{sun2016deep}. The major difference between these methods are in the design of their losses. DANN loss is an adversarial loss, while AssocDA and CORAL losses measure the assimilation distance of the common features from the two domains. Our implementation of these methods uses exactly the losses from the original papers. To encourage the extraction of the common features, we opt to set the dimensionality of the common feature space to be smaller than those in the authors' original implementation/report.

Note that the performance of our implementation differs than those reported from the original papers (see Figure~\ref{fig:accvis}). For DANN, our implementation performs significantly better than the authors' implementation.  For AssocDA, due to the use of a smaller dimensionality for the common feature space, the initial DA performance ($i=0$) is inferior to the authors' implementation. The performance of CORAL on MNIST $\rightarrow$ MNISTM is not reported in the original paper, but we found that the performance of our implementation is comparable with that reported in~\cite{bousmalis2016domain}.

\begin{table}[b!]
  \centering
  \begin{tabular}{c l l l}
    \toprule
    Method     & MNIST$\rightarrow$MNISTM     & MNIST$\rightarrow$USPS     & SVHN$\rightarrow$MNIST  \\
    \midrule
    Source-only & 52.25  & 78.9    & 54.9   \\
    \hline
    \hline
    CORAL \cite{sun2016deep}       & 52.30    & 81.7        & 63.1\\
    MMD \cite{tzeng2014deep}         & 76.9  & 81.1    &71.1      \\
    DANN \cite{ganin2016domain}        & 75.4  & 85.1    &73.85    \\
    DSN \cite{bousmalis2016domain}         & 83.2  &91.3        &82.7     \\
    AssocDA \cite{haeusser2017associative}     & 89.5  &89.6        &\textbf{97.6}\\%[0.01\linewidth]
    %Ours/no iteration        &86.8   &89.39    &82.9\\
    %\textbf{Ours} / CORAL        & 57.2   & 75.4    & 63.5 \\
    %\textbf{DIDA /wo}         & 86.8   & 89.39    & 82.9\\
    %\textbf{Ours} / AssocDA     &  91.5   & XX.X    & XX.X \\
    \hline
    DiDA (backbone)        & \textbf{92.9} (86.8)  & \textbf{92.5}  (89.39)  & 83.55 (82.9) \\
    \hline
    \hline
    Target-only  &97.8 &95.8 &99.42 \\
    \bottomrule
  \end{tabular}
    \caption{\label{tab:DA_performance}
    	Classification accuracy (\%) for several unsupervised domain adaptation tasks. The ``Source-only'' and ``Target-only'' rows report the classification accuracy on the target domain without domain adaptation after supervised training on the source domain or target domain, respectively. The performance of DANN backbone, as well as the performance of our method based on this backbone, together with the results from several unsupervised domain adaptation methods are reported in this table,  with the best performance highlighted in bold.}
\end{table}

Our method leverages the initial domain adaptation for disentangled synthesis (Di), then the synthetic target domain data are used for training the domain adaptation (DA) again. We show that the domain adaptation results can be improved over several such DiDA iterations in Figure~\ref{fig:accvis} (blue) on MNIST $\rightarrow$ MNISTM. In the same figure, we also plot for reference the domain adaption results with equal number of training epochs, but without updating training data (in orange). Clearly, the improvements are brought by DiDA iterations, but not by training the domain adaptation with more iterations. We show that DiDA brings 6.1\% improvement (86.8\% $\rightarrow$ 92.9 \%) over the already very strong DANN baseline. For AssocDA backbone, though the initial DA performance of our implementation is inferior to that of the authors' implementation, the final performance (92.5\%) surpasses the original AssocDA with a 2.9\% margin, where DiDA brings 19\% improvement (72.5\% $\rightarrow$ 91.5\%). For CORAL backbone, note that the training with more iterations without DiDA suffers from overfitting, while DiDA is able to bring (4.9\%) improvement (52.3\% $\rightarrow$ 57.2\%).

\paragraph{Evaluation on Different Tasks.}
We conducted quantitative evaluation of our method (with DANN backbone, unless otherwise noted) on three benchmark tasks, and summarized the results in Table~\ref{tab:DA_performance}, in comparison with the results from several state-of-the-art domain adaptation methods.

Our method performs better than other methods on both MNIST $\rightarrow$ MNISTM and MNIST $\rightarrow$ USPS tasks, with 6.1\% and 3.11\% improvement brought by the DiDA iterations over the backbone method, respectively. Note that these performances are quite close to the performance with supervised training in the target domains. Our method is able to boost the performance on SVHN $\rightarrow$ MNIST as well, however, with an improvement which is relatively smaller than those on the other two datasets. We suspect the reason is that the target domain, MNIST, contains minimal domain-specific features, thus the interference of the target domain specific features is small in this task. While disentangled synthesis can still improve over iterations, the benefit is less significant. Based on this experiment, we believe our method has more potential in boosting performance for domain adaptation scenarios where the target domains contain rich domain-specific features.

\begin{figure*}[t!]	
	\centering
	\centering
	\includegraphics[width=1.0\textwidth]{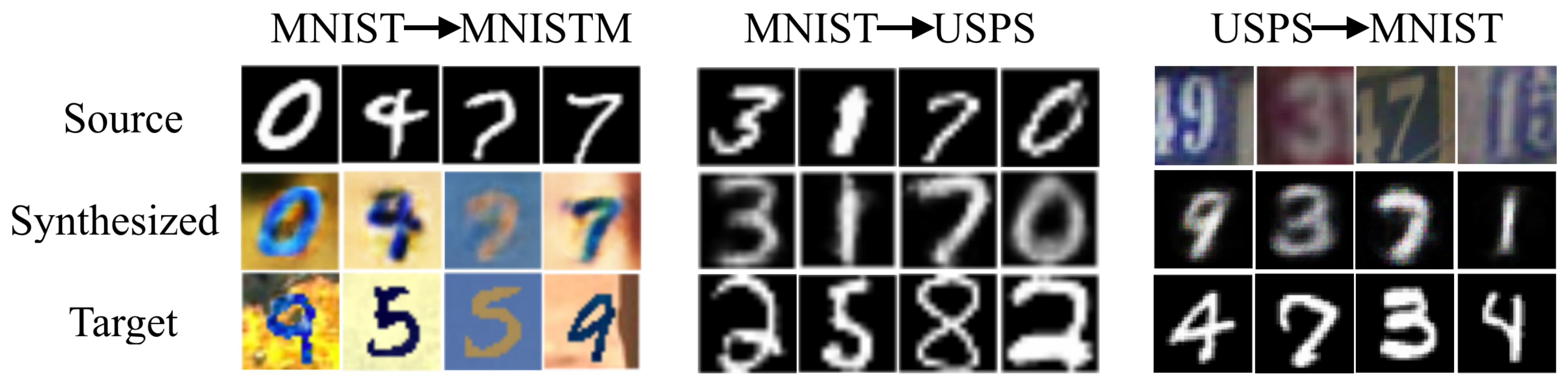}
	\vspace{-0.05\linewidth}
	\caption{Synthetic target domain images for different settings. The images in the top row are from the source domain, while the images in the bottom row are from the target domain. The synthetic target domain images in the middle row are generated by combining the common feature from the top row images and the domain specific features of the bottom row.}
	\label{fig:synth-domain}
\end{figure*}

\begin{table}[h!]
	\centering
	\begin{tabular}{c c c c c c c}
		\toprule
		Task   & \multicolumn{2}{c}{MNIST$\rightarrow$MNISTM} &\multicolumn{2}{c}{MNIST$\rightarrow$USPS}  &\multicolumn{2}{c}{SVHN$\rightarrow$MNIST}\\
		&common &specific &common &specific &common &specific \\
		\midrule
		Random chance    & 10 & 10 & 10 & 10 & 10 & 10  \\
		\hline
		DiDA   & 98  & 12  &95  & 11  &95  & 20\\
		\bottomrule
	\end{tabular}
	\caption{\label{tab:DI_performance}
		Comparison of classification accuracy (\%) using common features and domain-specific features extracted by our method. The common features can be leveraged by a classifier for accurate classification prediction, while the classifiers perform poorly on the domain-specific features.}
\end{table}

The common and domain-specific feature disentanglement is crucial for the successful boosting of domain adaptation methods. In addition to the evaluation of performance on the domain adaptation tasks, we quantitatively analyze the quality of common feature and specific feature disentanglement using the protocol described in \cite{mathieu2016disentangling}. We train a shallow neural network by taking the common feature and specific feature as input to predict labels of source and target domain datasets. The classification accuracy, reported in Table~\ref{tab:DI_performance}, shows that the common feature encodes the class information well, while the specific feature is agnostic to the classification task, as the accuracy based on them is close to that of random selection. The big gap between the classification accuracies indicates a high quality feature disentangling in our method, in supporting of the domain adaptation performance boosting.

\subsection{Qualitative Evaluation}

\begin{figure*}[t!]	
	\centering
	\centering
	\includegraphics[width=1.0\textwidth]{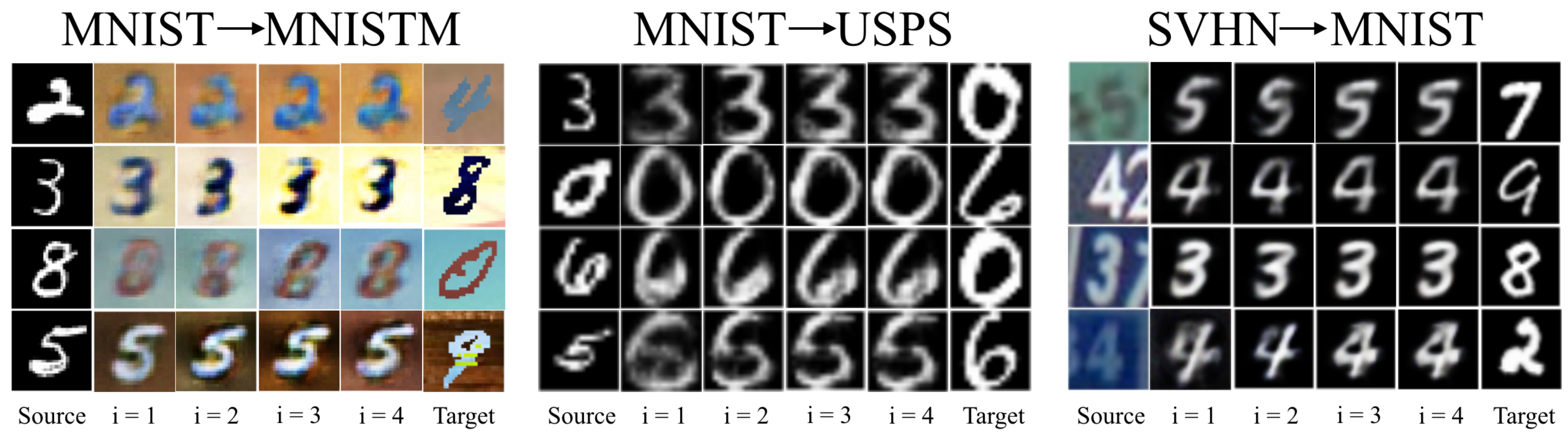}
	\vspace{-0.05\linewidth}
\caption{Synthetic target domain images over DiDA iterations. In each set, the image on the left side is the source image, from which we extract the common features, and the image on the right side is the target image, from which we take the specific features.}
\label{fig:itervis}
\end{figure*}

An important by-product of our method is the synthetic target domain images generated by combining common features from source domain and specific features from target domain. These images inherit the ground truth labels from source domain, while being visually close to the target domain images. The visual similarity between these synthetic and real target domain images can serve as a good qualitative indicator of the synthesis quality. We show some samples of the synthetic target domain images in Figure~\ref{fig:synth-domain}, together with the corresponding source domain images from which the source domain features are sampled and target domain images from which the target domain specific features are sampled. As can be seen from these images, the synthetic target domain images share the common ``content'' as the source domain images, while being visually similar to target domain images.

In Figure~\ref{fig:itervis}, we visualize the synthetic target domain images over DiDA iterations. Note that the quality of the synthetic target domain images improves over DiDA iterations. From Figure~\ref{fig:accvis}, we conclude that the domain adaptation is improved over DiDA iterations. Here, together with the qualitative improvement over the synthetic images, we can conclude that not only ``Di'' improves ``DA'', but also ``DA'' improves ``Di'' --- the disentangled image synthesis.

\begin{figure}[h!]
\centering
\begin{minipage}[t]{0.45\textwidth}
\centering
\centerline{Common features}
\begin{minipage}[t]{0.40\textwidth}
\centering
%\subfigure[{common feature, $i=1$}] { \label{fig:a}
\includegraphics[width=1.0\columnwidth]{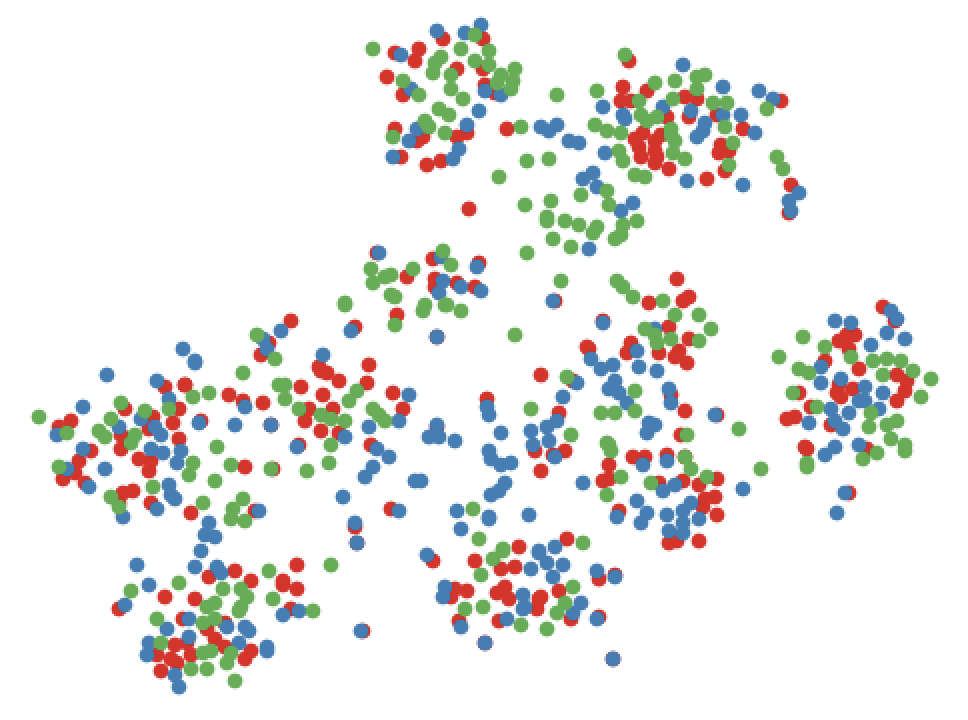}
\centerline{$i=1$}
\end{minipage}
\begin{minipage}[t]{0.40\textwidth}
\centering
\includegraphics[width=1.0\columnwidth]{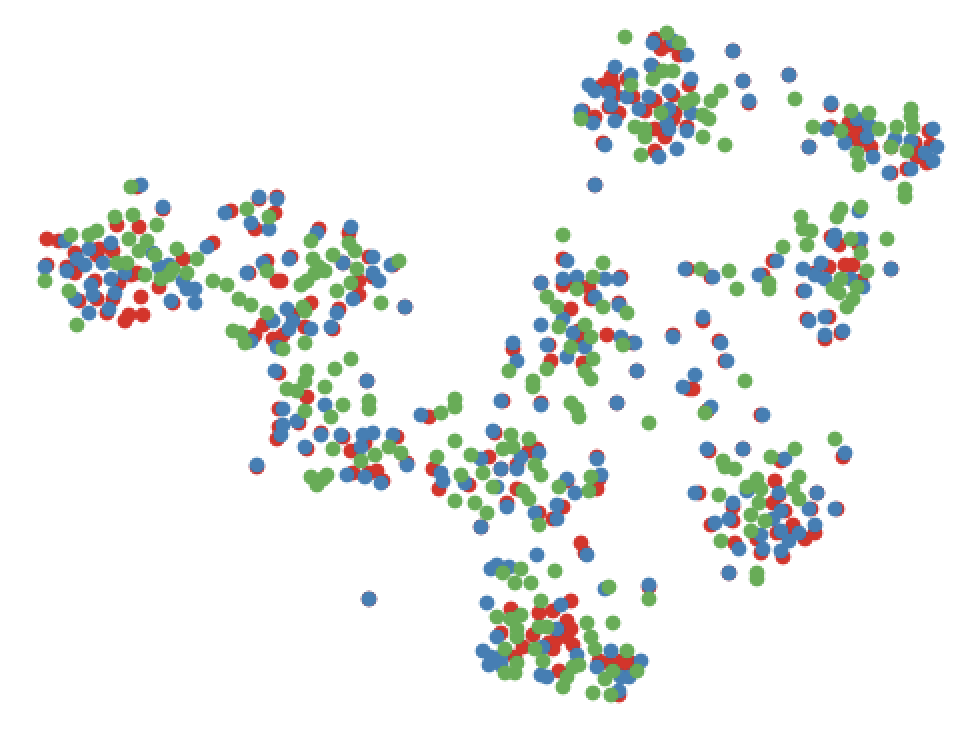}
\centerline{$i=4$}
\end{minipage}
\end{minipage}
\hspace{0.04\textwidth}
\begin{minipage}[t]{0.45\textwidth}
\centering
\centerline{Specific features}
\begin{minipage}[t]{0.40\textwidth}
\centering
\includegraphics[width=1.0\columnwidth]{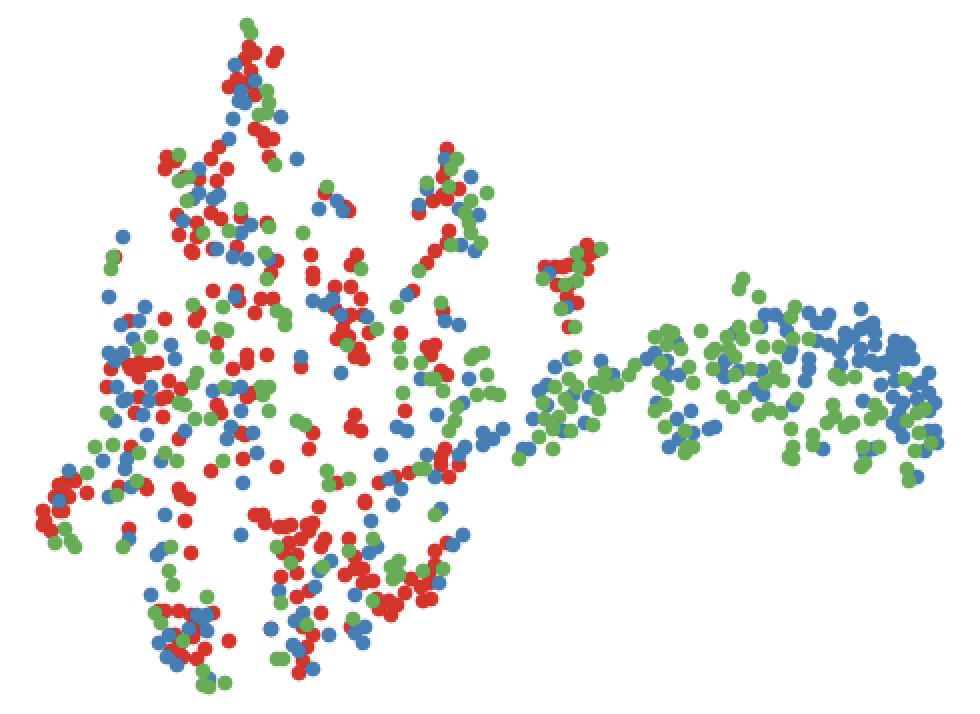}
\centerline{$i=1$}
\end{minipage}
\begin{minipage}[t]{0.40\textwidth}
\centering
\includegraphics[width=1.0\columnwidth]{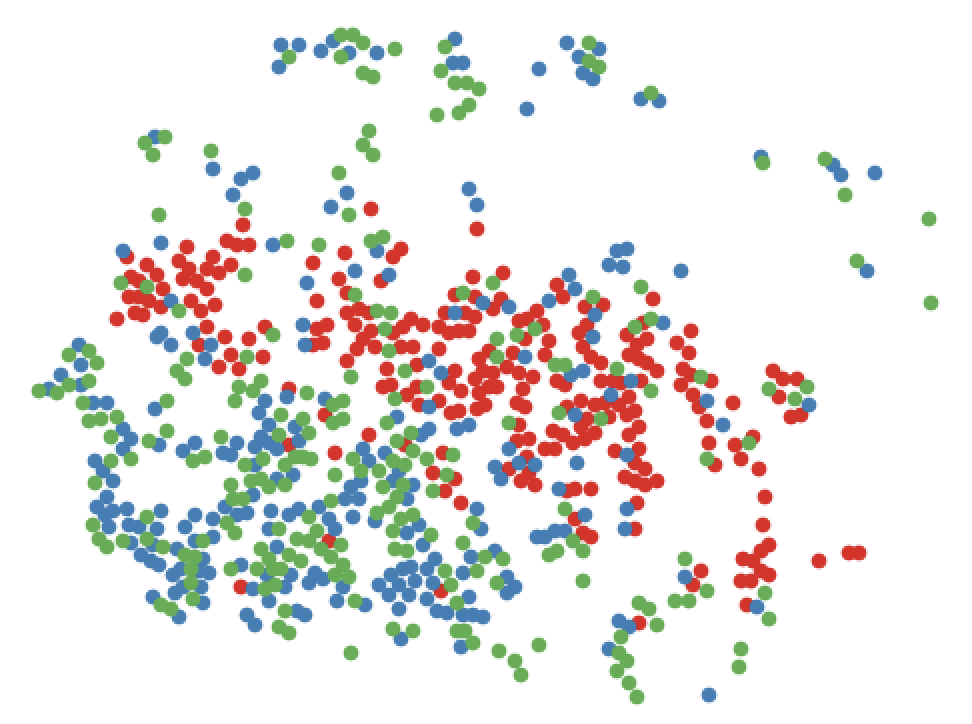}
\centerline{$i=4$}
\end{minipage}
\end{minipage}
\caption{t-SNE visualization of common and specific features of source (red), target (green) and synthetic target (blue) domain images after the first and fourth DiDA iterations, respectively.}
\label{fig:tsne}
\end{figure}

We also visualize the ``Di'' improvement in the feature space. The common features and specific features of source, target and synthetic target domain images after the first and fourth DiDA iterations are visualized by t-SNE in Figure~\ref{fig:tsne}. Note that the common features of the source and target domain images are more overlapped with each other, indicating the increase of their ``commonness''. They are also becoming more clustered over iterations, indicating the improvement in the classification accuracy. In contrast, the domain-specific features of the source and target images become more separated from each other over iterations, indicating that they
encode more specific information.

\section{Conclusion}
\label{sec:conclusion}

We proposed a method for boosting domain adaptation with disentanglement analysis. We show that the ``Di'' (disentangled synthesis) and ``DA'' (domain adaptation) can help each other over DiDA iterations, and eventually boost domain adaptation performance. Our method is a general framework that could potentially work with a variety of domain adaptation and feature disentanglement methods.
Further research is necessary to explore which specific combinations of domain adaptation and disentanglement methods would yield the best performance for different tasks.

%\subsubsection*{Acknowledgments}

%The work is supported in part by ...

\small
\begingroup
	\setlength{\bibsep}{1pt}
	\bibliographystyle{plain}
	\bibliography{references}
\endgroup

\end{document}